# Characterizing Perspective Error in Voxel-Based Lidar Scan Matching

Jason H. Rife  |  Matthew McDermott

Tufts University

**Correspondence**
Jason H. Rife
Department of Mechanical Engineering
Tufts University, Medford,
MA, 02155, USA.
Email: jason.rife@tufts.edu

**Abstract**

This paper quantifies an error source that limits the accuracy of lidar scan matching, particularly for voxel-based methods. Lidar scan matching, which is used in dead reckoning (also known as *lidar odometry*) and mapping, computes the rotation and translation that best align a pair of point clouds. Perspective errors occur when a scene is viewed from different angles, with different surfaces becoming visible or occluded from each viewpoint. To explain perspective anomalies observed in data, this paper models perspective errors for two objects representative of urban landscapes: a cylindrical column and a dual-wall corner. For each object, we provide an analytical model of the perspective error for voxel-based lidar scan matching. We then analyze how perspective errors accumulate as a lidar-equipped vehicle moves past these objects.

## 1 | INTRODUCTION

Lidar systems are data-rich perception sensors, which cast optical beams in known directions and compute the time of flight for the reflected signal. The time of flight can be converted to distance; thus, each return corresponds to a vector (range and direction) describing a point on some surface in the surrounding environment (or *scene*). By scanning over a set of directions, the lidar creates an image, called a *point cloud*, which represents the scene's geometry. It is often useful to align (or *register* or *scan match*) more than one point cloud. Applications of this scan matching process include creating a map, inferring relative motion over time, or performing both via simultaneous localization and mapping, as described by Thrun et al. (2005).

Existing scan-matching algorithms can generally be classified into one of three categories: feature-based methods, voxel-based methods, and direct machine-learning (ML) methods. Feature-based methods seek to correspond recognizable features from a pair of point clouds, based on either individual points, as in the well-known iterative closest point (ICP) algorithm (Segal et al., 2009; Yuan et al., 2023), or extracted features such as planes (Besl & McKay, 1992), cylinders (Hassani & Joerger, 2021), or corners in the case of the widely used lidar odometry and mapping (LOAM) algorithm (Shan & Englot, 2018; Zhang & Singh, 2014).

Hybrid feature-based methods also exist, which combine ML-based strategies with feature extraction, such as the recent work by Plachetka et al. (2021), which applies ML to create bounding boxes for objects identified in a scene. Voxel-based methods, such as the normal distribution transform (NDT) and the iterative closest ellipsoidal transform (ICET) algorithms, divide the point cloud into volume elements (or *voxels*), each containing a distribution of points; correspondence is performed purely geometrically by attempting to align point distributions within each voxel (Biber & Straßer, 2003; Stoyanov et al., 2012; McDermott & Rife, 2022b; McDermott & Rife, 2023). Direct ML algorithms, such as those of Li et al. (2019), do not explicitly form correspondences or compute geometric transformations to relate point clouds; instead, they train neural networks to align point clouds directly.

In our work with urban lidar data sets, we have observed similar levels of accuracy for a wide range of scan-matching algorithms, including ICP, LOAM, NDT, and ICET, with different algorithms delivering slightly more accuracy in different types of urban and suburban scenes. Given their similar levels of accuracy, these algorithms are primarily distinguished by how they fail. In our work, we have placed a particular emphasis on voxel-based algorithms because we believe that they are well suited to high-integrity, safety-critical systems, such as automated ground transportation.

For safety-of-life navigation systems, voxel-based algorithms offer the benefits of being both interpretable and free of data-association ambiguities[1]. By contrast, feature-based algorithms may extract different feature sets from each frame; thus, it may not be possible to find matches for features between time steps. Consequently, incorrect-fix probabilities must be computed, or alternatively, features must be purposefully engineered for reliable extraction (Hassani et al., 2018; Nagai et al., 2023). Moreover, ML-based scan matching is opaque. Although efforts are underway to characterize errors in ML scan matching (Joerger et al., 2022), ML methods still introduce the risk of rare unpredicted faults (Willers et al., 2020), which can make integrity arguments difficult.

Recognizing that voxel-based algorithms offer some advantages, it is important to note that they also have limitations. Most notably, voxel-based algorithms appear to be more susceptible than other approaches to perspective errors, sometimes referred to as *self-occlusions* (Xu et al., 2022). Perspective errors occur because of missing data. As a lidar moves past a static three-dimensional (3D) object, previously hidden patches of the object may appear or disappear because of the object's shape. This phenomenon is visible in the Ouster lidar data shown in Figure 1, where driving around the corner of a building reveals missing data along a previously hidden wall. The data illustrated were obtained from the manufacturer's website for the OS1 128-channel sensor (Ouster, 2023). Whereas objects (such as the building) may partially occlude themselves, they may also occlude each other, a phenomenon known as *shadowing*. Evident in Figure 1, where tall posts "cast a shadow" on the ground and walls behind, this phenomenon has been addressed elsewhere (Hassani & Joerger, 2021; McDermott & Rife, 2022a) and is distinguished from perspective effects in that shadows fall along radial spokes traced outward from the lidar unit.

No prior efforts have been made to analyze perspective effects on scan matching. Thus, the main contribution of this paper is to develop a theoretical basis for quantifying perspective errors for voxel-based scan matching.

---

[1]Exceptions include the case of aliasing, in which the same subscene is repeated (as in a long colonnade), or the case of degenerate scenes (as in a completely flat field); such cases are equally problematic for all scan-matching algorithms.

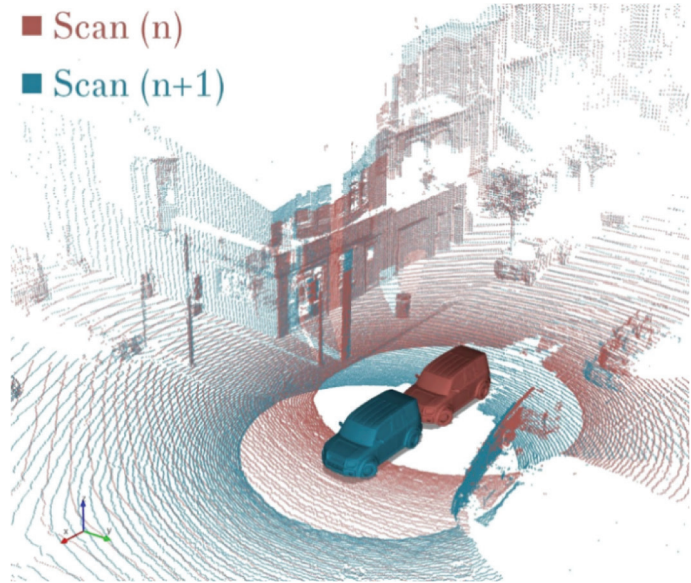

**FIGURE 1** Missing data appear as a vehicle drives around a corner. Two sequential point clouds are shown, with both generated by the same vehicle, equipped with an Ouster OS1 128-channel sensor. The second point cloud (blue) reveals a wall around the corner of the building (left side of image), a wall that was not visible in the first point cloud (red).

In the next section, we motivate this problem in qualitative terms. Subsequently, with the goal of obtaining representative bounds for perspective errors, we analyze the worst-case situation in which the self-occluding object is fully contained within a single voxel. Specifically, we analyze two representative types of self-occluding objects: a cylindrical column and a dual-wall corner. For these objects, we derive analytical equations to model perspective errors. We then interpret these analytical models using simulations, with the goal of informing mitigation strategies to minimize the impact of perspective errors in future lidar applications.

## 2 | QUALITATIVE MOTIVATION

Before analyzing perspective errors using mathematical analysis, it is helpful to first motivate the problem by discussing perspective errors in qualitative terms. This section aims to motivate the problem by providing evidence for perspective errors in lidar odometry data, briefly describing how algorithms obtain a pose solution from voxel-based measurements, discussing how missing data manifest on curved surfaces, and examining how lidar cloud-point density is related to the design of the lidar unit.

### 2.1 | Evidence for Perspective Errors in Lidar Odometry

The field of uncertainty quantification for lidar scan matching is relatively new, and the primary focus has been on quantifying random errors. Uncertainty quantification for voxel-based scan matching (Stoyanov et al., 2012; Hassani & Joerger, 2021; McDermott & Rife, 2022a) can generate representative error bounds; however, occasional systematic biases may violate the assumptions used to define these bounds. An important element of developing a case for algorithm integrity is to analyze these types of uncommon systematic bias.

An example of systematic error in lidar odometry is shown in Figure 2. Lidar odometry is the process of computing a relative position over time by cascading a series of transforms (translation and rotation) derived from a series of scan matches. In Figure 2, ICET (McDermott & Rife, 2023) is applied to scan-match lidar point clouds drawn from the KITTI data set (Geiger et al., 2013). The scan-match results are accumulated to generate a lidar odometry signal and referenced to Global Positioning System (GPS)/inertial navigation system (INS) ground truth to define an error signal. The error signal is plotted in the figure as a dark line, atop a shaded region that represents the $2\sigma$ bound accounting for the random error of the combined sig-nals. The shaded region contains the error signal (as expected) for the first t = 0–11 s, beyond which a measurement bias captures the system, introducing an additional 0.5 m of error by t = 14 s. In this case, the ICET processing contains mitigation to account for extended-object, ground-plane, and shadowing errors (McDermott & Rife, 2023). The car is moving into an intersection at this point; hence, possible remaining sources for systematic error include motion distortion (or *rolling shutter*) and perspective effects. Mathematical analyses of perspective and motion-distortion effects are both needed to disentangle how the effects combine in this situation; mathematical analysis is also needed to inflate the bound appropriately (to contain the error signal) or to mitigate errors (to keep the error signal inside the existing bound). Although analysis is needed to understand both effects, we focus here on perspective errors and leave an analysis of motion-distortion effects to future work.

## 2.2 | Internal Function of Voxel-Based Scan-Matching Algorithms

The relative position updates computed at each time step in Figure 2 are the outputs of a voxel-based scan-matching algorithm, which maps voxel-level measurements to a pose transformation (a combined translation and rotation) of the lidar unit between successive frames. The main goal of this paper is to examine perspective effects at the measurement level, within one voxel. To determine how the pose solution relates to individual voxel-level measurements, it is helpful to briefly review how voxel-based scan matching works. Figure 3 provides a simple illustration of the process. Initially, two point clouds (shown as collections of red and blue dots in the figure) are aligned via an initial guess. If a sufficient number of points are present from both clouds within any given voxel, then a mean and covariance ellipse can be computed for the points within that voxel. An individual voxel-level measurement consists of the vector difference between the two point-cloud means. The pose transform is then determined as the operation that minimizes the magnitude of the mean-difference vectors over all voxels. The error output from the process can be described in terms of the random noise and systematic biases for each voxel measurement. The solution process weights each voxel measurement by its uncertainty to compute the solution; the uncertainty in the solution can be characterized by mapping the random noise and biases through the solution process, which is a standard least-squares solution for the ICET algorithm (McDermott & Rife, 2022b; McDermott & Rife, 2023).

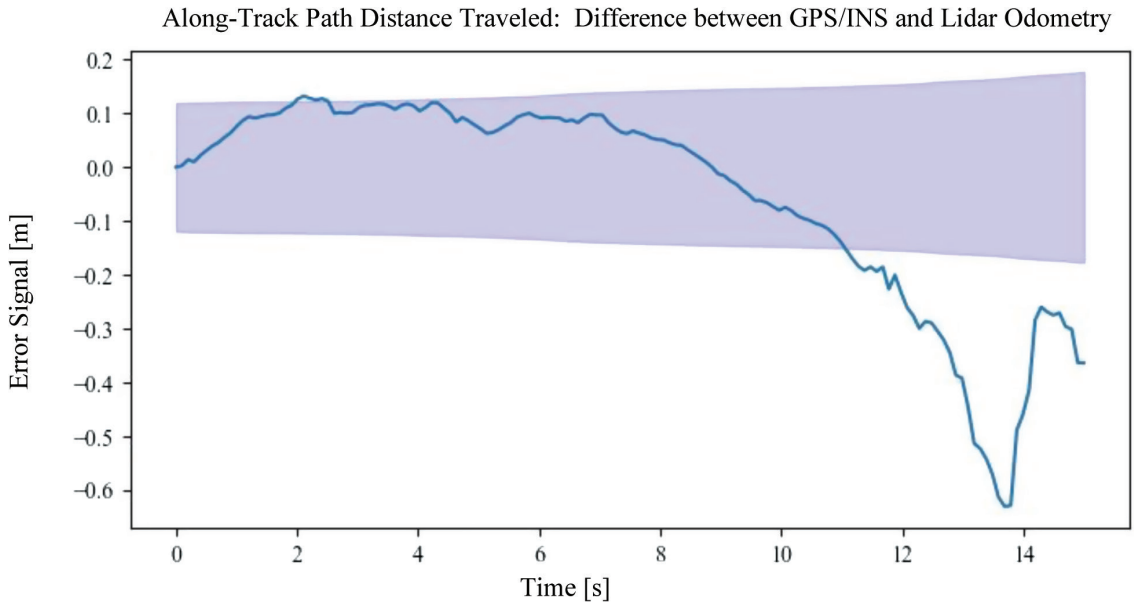

**FIGURE 2** When voxel-based algorithms such as NDT or ICET are used in lidar odometry, spiking errors may appear near corners. In this case, the ICET algorithm (McDermott & Rife, 2022b) was used to perform scan matching and uncertainty quantification for KITTI data record "9/26 drive 005" (Geiger et al., 2013). Lidar odometry is referenced to a GPS/INS ground truth to obtain an error signal (dark line), which is plotted with a 95% confidence limit (lightly shaded region). The error spikes outward from the confidence bound at t = 12 s.

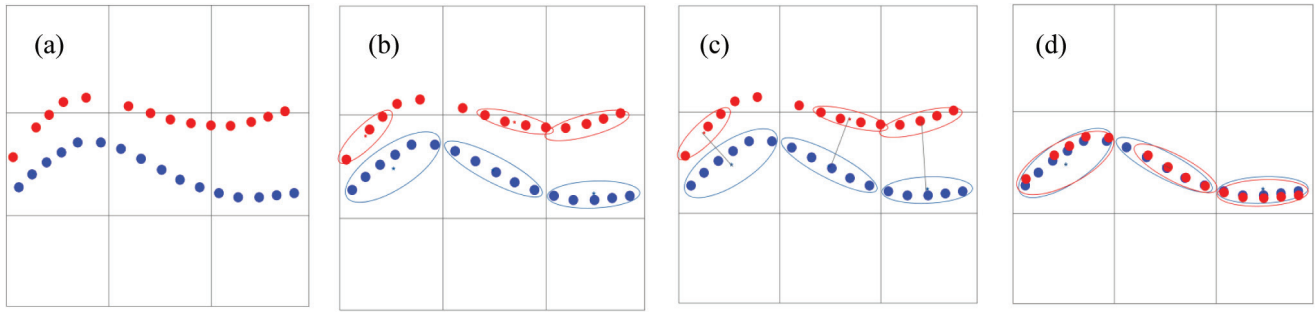

**FIGURE 3** The process of voxel-based scan matching
Two point clouds (red and blue) are compared. (a) The points in each cloud are associated with a particular voxel in a regular grid. (b) For voxels in which both of the point clouds are represented, a covariance ellipse is fit to each set of points. The center of each ellipse corresponds to the mean of the associated points. (c) The offset between the mean values is computed within each voxel. (d) A rigid transformation is then determined to rotate and translate the second (red) point cloud, to best align it with the first (blue) point cloud.

In our subsequent analysis, we are interested in voxel-level biases introduced by perspective error; these measurement biases propagate through the solution process, resulting in biases in the resulting pose solution.

## 2.3 | Perspective Error on Curved Surfaces

To evaluate the measurement biases that may be caused by perspective errors within a single voxel, we consider the worst-case situation with an object of interest contained entirely within one voxel. (Splitting the object across many voxels reduces perspective errors, as discussed in Section 5.) The magnitude of the measurement bias is dependent on the geometry of the self-occluding object. To this end, we consider two types of objects: a corner (with a discontinuous surface) and a cylinder (with a rounded and continuous surface). Whereas perspective errors trigger suddenly when moving around a corner (as illustrated by the sudden appearance of new points in Figure 1), perspective errors accumulate somewhat more gradually for curved surfaces.

To illustrate the differences between sharp corners and rounded surfaces, consider Figure 4, which shows how missing data can arise depending on the location of the lidar unit. This figure presents a top-down view of a cylindrical column. In Figure 4(a), the column is viewed from two lidar locations (crosses) that are separated laterally. The point clouds generated from each location share a common overlapping region (green arc), but each point cloud also includes samples from a non-overlapping region (red arcs). The "missing data" from each point cloud lead to a difference in the mean location of the samples in each case (with the distinct point-cloud mean locations each identified by a star). Algorithms such as the NDT (Biber & Straßer, 2003) and ICET (McDermott & Rife, 2022b) attempt to align the point-cloud means between two scans. Assuming that the column is an immobile object, then aligning the two means (stars) results in the false inference that the two viewpoints (crosses) are closer together than their true distance.

Perspective errors are generally less severe for radial motion than for lateral motion, as shown in Figure 4(b). Although the symbols in Figure 4(b) match those of the rest of the figure, the lidar locations are different. In Figure 4(b), the two lidar locations are aligned on the same spoke radiating from the cylindrical column's centerline. Again, there is an overlapping set of samples (on the green arc) and a non-overlapping set of samples (red arc); however, in this case, the more distant lidar scan (blue) visualizes the entire region seen by the nearer scan (brown). Importantly the data missing from the nearer scan (red arcs) are symmetric; consequently, the point-cloud means (stars) are offset only in radius and not in circumferential coordinates.

These simple examples demonstrate that even a smoothly curved, convex surface (such as a cylindrical column) can occlude the lidar's view and corrupt the resulting point cloud with missing data. Even more dramatic occlusions occur for more complex objects, such as those involving sharp corners or outcroppings. In this paper, we do not consider signal-miss effects due to reflective surfaces (Xu et al., 2022).

Noting that the properties of curved surfaces are somewhat different from those of discontinuous surfaces, this paper focuses on analyzing one characteristic object from each category: a cylindrical column and a dual-wall corner.

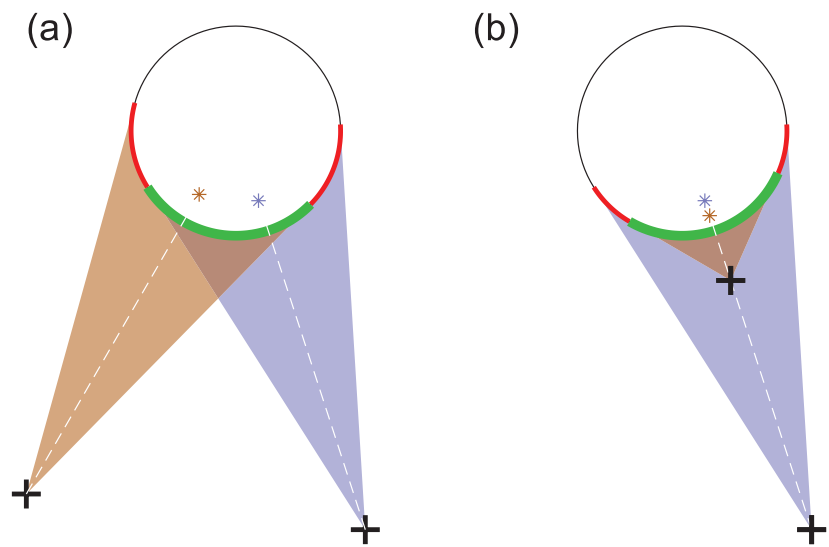

**FIGURE 4** Changes in perspective for (a) lateral and (b) radial movement of the lidar between two locations (cross markers)

## 2.4 | Modeling Point-Cloud Density

In analyzing perspective errors for an object such as a column or corner, we must model the point-cloud density over the object's surface. Lidar samples are not uniformly separated in object-centered coordinates (e.g., they are not spaced at equal distances along a flat wall). Rather, for a rotating lidar unit, samples are spaced at roughly even angular increments in azimuth (and at fixed angular increments in elevation, which are not necessarily equal). For a non-rotating solid-state (or "flash") lidar unit, the sample density is based on perspective geometry, similar to the case for a camera. In this paper, we focus on rotating lidar units, which are commonly used for automotive applications (in large part because static "flash" lidar units have not achieved the resolution needed for precision localization).

Assuming a rotating lidar unit, we assume that lidar samples are spaced with uniform density in azimuth angle. The result is visualized for a two-dimensional (2D) scan of a wall, as seen from above, in Figure 5. For this scan (assuming a zero azimuth angle), the angular density is uniform, but the point density along the wall changes when the lidar is moved between two positions (marked by crosses in Figure 5(a) and Figure 5(b)). There are no missing data because the lidar views the entire face of the wall from any viewing location (unless it moves behind the wall). However, the "center of mass" of the point cloud moves slightly because of density effects. Let us consider the difference in the distribution of samples moving from a viewing location on the left of the wall (Figure 5(a)) to a more central viewing axis (Figure 5(b)). When viewed from the left, the point density is much higher on the left than the right. When viewed from the middle, the point density is highest in the middle. For the location at which the point density is the highest, the interval between samples is the shortest (as shown in Figure 5(c)). In this particular case, the high density of samples on the left side of the wall in Figure 5(a) will shift the point-cloud mean slightly to the left as compared with Figure 5(b). Because these density effects shift the point-cloud distribution, we consider point density as well as self-occlusion effects in our subsequent analysis of perspective errors.

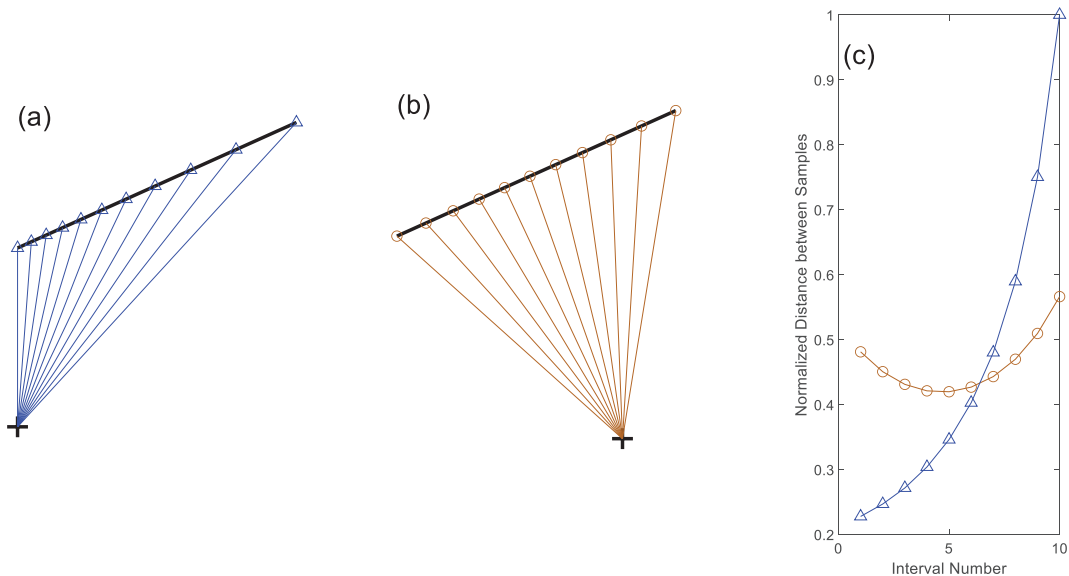

**FIGURE 5** Point density along a planar surface as a function of lidar location
A short wall is shown, visualized from two locations indicated by cross markers. When the lidar is located at the left cross, points are sampled as shown in (a), and when the lidar is located at the right cross, points are sampled as shown in (b). Distances between sequential points as a function of point index (counting from left to right) are plotted in (c).

In the next section, we analyze perspective effects in 2D (on the zero-elevation plane) with the key assumption that objects are fully contained within one voxel. We consider this assumption to be the worst-case scenario in the sense that higher-resolution grids (which subdivide objects into smaller point clouds) reduce curvature and reduce perspective errors, as we will discuss in Section 5.

## 3 | ANALYTICAL MODELS FOR THE POINT-CLOUD MEAN

This section develops analytical models to quantify perspective errors associated with two types of objects: cylindrical columns and dual-wall corners. These objects are representative of features common in urban environments. Because we focus on voxel-based algorithms that perform scan matching by attempting to align the point distributions within each voxel, we assess the systematic bias as equal to the shift in the point cloud's mean location. We assume dense, effectively continuous, sampling and assume that the point cloud covers the entire surface of the object that is visible from the lidar's instantaneous location. As a further simplification, we assume that the object is located entirely inside a single voxel. This is not an unreasonable assumption, as scan-matching algorithms often use voxels with a span of 1 m or more.

### 3.1 | Perspective Error for a Cylinder

A top-view schematic for analyzing lidar scans of a cylindrical column is shown in Figure 6. In this schematic, the lidar is located at point L (cross marker), and the centerline of the cylinder passes through point O. The sample point *S* lies on the surface of the cylinder, which has a radius *R*. The point cloud is the locus of all visible points *S* along the cylinder's surface (green highlighted arc in Figure 6).

Our approach for computing the mean location of the point cloud is to integrate the coordinates *x* and *y* along the appropriate cylindrical arc (between the brown dashed lines in Figure 6). The *x* and *y* Cartesian coordinates are defined by the unit vectors $\hat{\mathbf{u}}_x$ and $\hat{\mathbf{u}}_y$ in the diagram. Although the cylinder and point cloud are defined in 3D, we can compute the mean location for the point cloud using a 2D analysis, because the cylinder has a constant cross-section out of the plane. (Assuming that points are evenly distributed in the vertical direction, the vertical component of the point-cloud mean lies at half the height of the cylinder.) Our primary goal is to compute the change in the observed point-cloud mean when the cylinder is viewed from two different points *L*.

In computing the shift in the mean location of the point cloud, we consider not only missing data (as shown in Figure 4), but also point-density effects (as shown in Figure 5). To this end, we model the point-cloud density $\rho$ as uniform in the angular coordinate $\theta$ about the lidar location *L*. Modeling the density as a function of $\theta$ (rather than, say, of the Cartesian coordinates *x* and *y*) automatically accounts for point-density variations along object surfaces, as shown in Figure 5.

The number of points *N* sampled across the cylinder arc can be obtained by integrating the density. With the point *L* defined as the origin of a cylindrical coordinate system, the point *O* lies at $(r_0, \theta_0)$. The upper and lower bounds (brown dashed lines in Figure 6) are displaced from $\theta_0$ by counterclockwise-positive angles $\theta_U$ and $\theta_L$, noting that $\theta_L = -\theta_U$ because of the symmetry of the cylinder. Without a loss of generality, we can redefine the circumferential angle relative to $\theta_0$ (i.e., set $\theta_0 = 0$) and thus obtain the following:

$$N = \int_{\theta_L}^{\theta_U} \rho(\theta) d\theta \qquad (1)$$

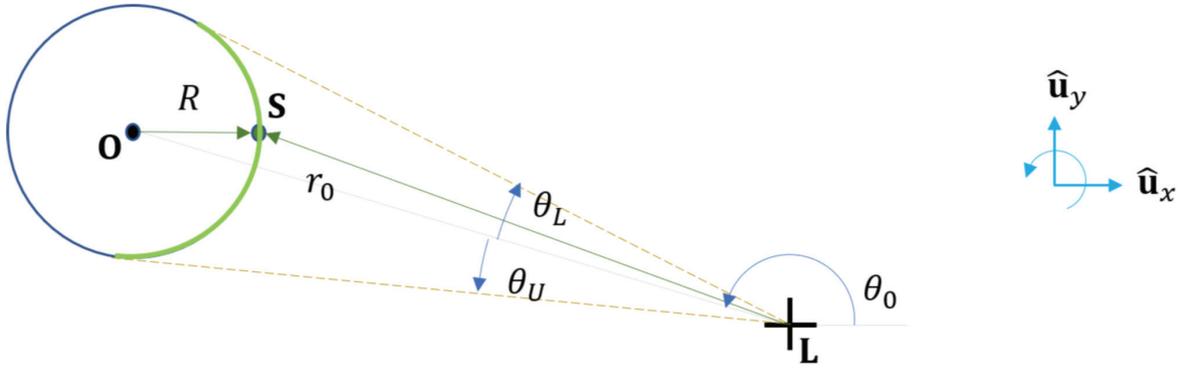

**FIGURE 6** Top view of a cylindrical column

Noting that the density is uniform, or constant, in the angular coordinate, we can factor out the density and evaluate the integral to give the following:

$$N = \rho(\theta_U - \theta_L) \qquad (2)$$

The mean location over all points in the point cloud, sampled at angles $\theta_i$, is given as follows:

$$\bar{\mathbf{x}} = \frac{1}{N} \sum_i \mathbf{x}_S(\theta_i) \qquad (3)$$

Each sample $\mathbf{x}_S$ comprises a nominal value and a random measurement error; the linear structure of Equation (3) indicates that the random errors and systematic biases in the nominal value superpose and can therefore be modeled separately. Because we have analyzed the random error elsewhere (McDermott & Rife, 2023), we focus only on the deterministic bias in this paper.

If the points are sampled with sufficient density, we can replace the summation in Equation (3) with an integral:

$$\bar{\mathbf{x}} = \frac{\int_{\theta_L}^{\theta_U} \mathbf{x}_S(\theta) \rho(\theta) d\theta}{N} \qquad (4)$$

By substituting Equation (2) into Equation (4) and noting that the density can again be factored out of the integral, we can simplify to obtain the following:

$$\bar{\mathbf{x}} = \frac{\int_{\theta_L}^{\theta_U} \mathbf{x}_S(\theta) d\theta}{\theta_U - \theta_L} \qquad (5)$$

The next step is to evaluate the integral in the numerator. Equation (5) uses a 2D vector notation, which will be important for the dual-wall corner; however, we can use a simpler scalar analysis for the cylinder, noting that the geometry is symmetric about the axis $\overline{LO}$. To determine the point-cloud mean for the cylinder, we need only solve for the component in the direction of $\overline{LO}$ (the x-direction for $\theta_0 = 0$). Using the coordinate system aligned with $\overline{LO}$, we thus have $\overline{\mathbf{x}} = [\overline{x}\ \ 0]^T$. Substituting this expression and invoking symmetry, we convert the general vector form of Equation (5) into a scalar equation for the cylinder:

$$\overline{x} = \frac{1}{\theta_U} \int_0^{\theta_U} x_s(\theta) d\theta \tag{6}$$

Using geometry and noting that the edges $\overline{LO}$ and $\overline{OS}$ have lengths $r_0$ and $R$, respectively, we can obtain the following:

$$(1 + \tan^2(\theta))x_s^2 - 2r_0 x_s + (r_0^2 - R^2) = 0 \tag{7}$$

Solving for $x$ and invoking standard trigonometric identities, we can solve this expression for $x_s$:

$$x_s(\theta) = r_0 \cos^2(\theta) - \cos(\theta)\sqrt{R^2 - \sin^2(\theta) r_0^2} \tag{8}$$

Only the negative root of the quadratic equation is considered here, because the positive root corresponds to the far (hidden) side of the cylinder. The analytical solution for Equation (6), with $x_s$ eliminated via Equation (8), is as follows:

$$\overline{x} = \frac{r_0}{2} + \frac{1}{4\theta_U}\left(r_0 \sin(2\theta_U) - \sin(\theta_U)\sqrt{2r_0^2 \cos(2\theta_U) - 2r_0^2 + 4R^2}\right)$$
$$- \frac{R^2}{2\theta_U r_0} \tan^{-1}\left(\frac{\sqrt{2}\sin(\theta_U)}{\sqrt{\cos(2\theta_U) - 1 + 2R^2/r_0^2}}\right) \tag{9}$$

Drawing a right triangle connecting $\overline{LO}$ to the tangent point from $L$, it is possible to show that the limit angle $\theta_U = \operatorname{asin}(R/r_0)$. Using the geometry of that right triangle, $R$ and $r_0$ can be largely eliminated from Equation (9) to give the following:

$$\overline{x} = \frac{r_0}{2}\left(1 + \frac{\sin(\theta_U)}{\theta_U}\left(\cos(\theta_U) - \frac{\pi}{2}\sin(\theta_U)\right)\right) \tag{10}$$

This value describes the distance from point $L$ to the mean along the direction of $\overline{LO}$. Subtracting Equation (10) from the length of $\overline{LO}$, which is $r_0$, and normalizing by the cylinder radius $R$, we can plot the point-cloud mean location relative to $O$, as shown in Figure 7. When the distance from the lidar to the cylinder center is at its minimum (e.g., when the lidar touches the cylinder, with $r_0 = R$), the lidar views only one point on the cylinder; thus, the point-cloud mean is located one radial unit from the center (as shown on the far left of Figure 7). As the lidar moves increasingly away from the cylinder (moving right on the horizontal axis of Figure 7), the distance from the cylinder center to $\overline{x}$ decreases, with the distance eventually reaching an asymptote (equal to $\frac{\pi}{4}$ or approximately 0.79 radial units) as the lidar distance grows toward infinity. It is worth noting, however, that because of the finite sample density of real-world scanning lidar units, the experimental performance at longer distances will not precisely match the theoretical performance, as we have assumed dense sampling at all distances.

If two lidar positions are considered, then we can compute the distance between the point-cloud mean vectors in each case. If a scan-matching algorithm aligns the two mean values, then we would obtain the systematic position-estimation bias for the voxel containing the cylinder. In differencing the point-cloud mean vectors, we must account for the different pointing direction of the vector (or equivalently, the distinct $\theta_0$ value) for each location. Using the index $j$ to distinguish the two lidar locations, we can write the vector from the point-cloud mean to the cylinder center $\mathbf{x}_O$ as follows:

$$\mathbf{x}_O - {}^{(j)}\mathbf{x}_s = \left({}^{(j)}r_0 - {}^{(j)}\bar{x}\right)\begin{bmatrix} \cos{}^{(j)}\theta_0 & \sin{}^{(j)}\theta_0 \end{bmatrix}^T \tag{11}$$

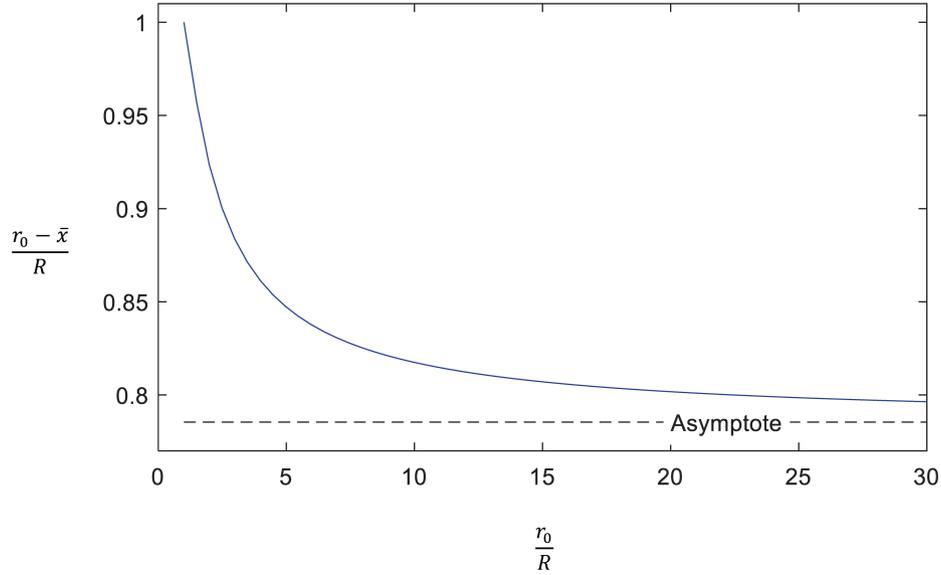

**FIGURE 7** Distance from point $O$ to the point-cloud mean in the direction of $\overline{LO}$

Using the cylinder center as a common reference, we can take the difference of Equation (11) for the two viewpoints $j$ to obtain the following:

$$\begin{aligned}{}^{(2)}\mathbf{x}_s - {}^{(1)}\mathbf{x}_s = &\left({}^{(1)}r_0 - {}^{(1)}\bar{x}\right)\begin{bmatrix} \cos{}^{(1)}\theta_0 & \sin{}^{(1)}\theta_0 \end{bmatrix}^T \\ &- \left({}^{(2)}r_0 - {}^{(2)}\bar{x}\right)\begin{bmatrix} \cos{}^{(2)}\theta_0 & \sin{}^{(2)}\theta_0 \end{bmatrix}^T\end{aligned} \tag{12}$$

The false movement inferred will be opposite the change in the point-cloud mean location; thus, the perspective error for the inferred lidar translation is $\epsilon = -\left({}^{(2)}\mathbf{x}_s - {}^{(1)}\mathbf{x}_s\right)$.

## 3.2 | Perspective Error for a Dual-Wall Corner

For comparison with the smooth cylindrical column, we also consider a second object with a sharp bend: the dual-wall corner shown from above in Figure 8. The corner consists of two thin walls of length $R$ that meet at vertex $O$. The lidar location is again labeled $L$. Each wall is oriented at a specified angle as viewed from above. Relative to $\overline{LO}$, the first wall is oriented at an angle $\psi_1$ and the second at $\psi_2$. Both angles are positive counterclockwise (so $\psi_1$ is negative as drawn).

The lidar detects points along the dual-wall corner, through a range of angles between a lower and upper bound ($\theta_L$ and $\theta_U$, respectively). The point cloud is generated from the entire visible surface, along both wall segments, each labeled with a subscript $i$. To determine the mean location for the point cloud, the first step is to write the location of an arbitrary point $S$ on one of the segments. We use a coordinate system in which the corner's vertex $O$ lies on the $x$-axis ($\theta_0 = 0$ for integration), at a distance of $r_0$ from $L$. In this configuration, the locus of points $S$ on one wall segment can be described by the following vector, where $R$ is the segment length, identical for both segments, and $f_i \in [0,1]$ is a fractional distance of $S$ along segment $i$. By letting $f_i$ vary over this range, we define the locus of all possible points along a wall segment, with the $f_i = 0$ case corresponding to the wall vertex at point $O$, the $f_i = 1$ case corresponding to the tip of the wall segment, and the $f_i = 0.5$ case corresponding to the midpoint of the wall segment:

$$\mathbf{x}_s = \mathbf{x}_O + \begin{bmatrix} r_0 + f_i R \cos(\psi_i) \\ f_i R \sin(\psi_i) \end{bmatrix} \tag{13}$$

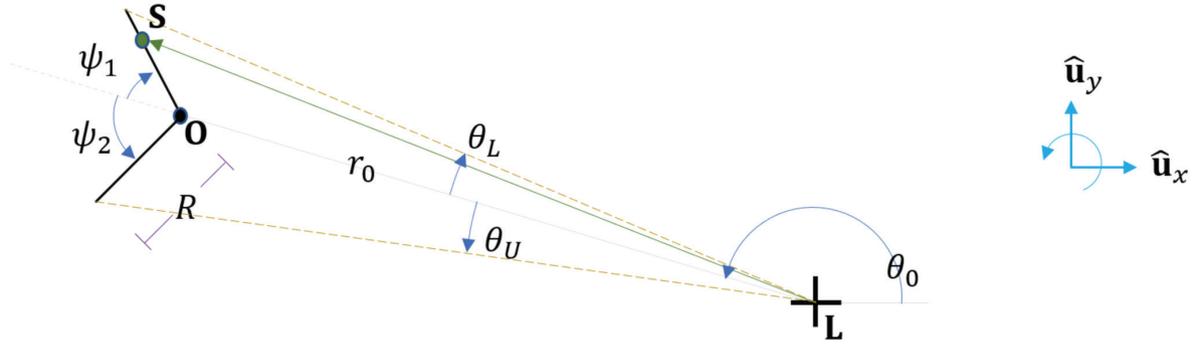

**FIGURE 8** Top view of the dual-wall corner

The point-cloud mean $\bar{\mathbf{x}}$ can be computed from Equation (5), where $\mathbf{x}_s$ is defined by Equation (13). Before performing the integration, however, it is first necessary to rewrite Equation (13) in terms of the angle $\theta$. The angle $\theta$ describes the direction of the vector $\mathbf{x}_s$; thus, we have the following:

$$\tan(\theta) = \frac{f_i R \sin(\psi_i)}{r_0 + f_i R \cos(\psi_i)} \tag{14}$$

Solving Equation (14) for the fractional length $f_i$ and substituting the result into Equation (13) gives the following expression:

$$\mathbf{x}_s = \mathbf{x}_O + r_0 \frac{\tan(\theta_i)}{\sin(\psi_i) - \cos(\psi_i)\tan(\theta_i)} \mathbf{u}_i \tag{15}$$

Here, we define the vector $\mathbf{x}_O$ to describe the position of $O$ relative to $L$ and the unit vector $\mathbf{u}_i$ to describe the direction from $O$ to $S$ (along the segment). The integral of $\mathbf{x}_s$ along wall $i$ depends only on the factor multiplying $\mathbf{u}_i$. For wall $i$, we define this integral, over the set of angles $\theta_i \in \Theta_i$, to be as follows:

$$g_i = r_0 \int_{\Theta_i} \frac{\tan(\theta_i)}{\sin(\psi_i) - \cos(\psi_i)\tan(\theta_i)} d\theta \tag{16}$$

We can now use Equation (5) to obtain the mean for the point cloud associated with the visible surface of the corner, by substituting Equations (15) and (16) into Equation (5). Considering both wall segments and modeling the walls as thin (so there is no additional contribution from the wall ends), the integral in Equation (5) can be written as follows:

$$\bar{\mathbf{x}} = \frac{g_1 \mathbf{u}_1 + g_2 \mathbf{u}_2}{\theta_U - \theta_L} + \mathbf{x}_O \tag{17}$$

To compare the shift in mean as viewed from two lidar locations, we need to introduce an index *j* to distinguish between the two lidar viewing locations. Introducing the location index via a leading superscript *j*, we can rewrite Equation (17) as follows:

$$^{(j)}\overline{\mathbf{x}} = \frac{^{(j)}g_1\,^{(j)}\mathbf{u}_1 + ^{(j)}g_2\,^{(j)}\mathbf{u}_2}{^{(j)}\theta_U - ^{(j)}\theta_L} + \mathbf{x}_O \tag{18}$$

The change in location of the point-cloud mean as viewed from the two lidar positions can be expressed by differencing Equation (18) for each of the two viewing locations *j*. In computing this difference, it is important to express the wall-tangent vectors $^{(j)}\mathbf{u}_i$ in a common coordinate system. To this end, we rotate from $\overline{LO}$-aligned coordinates back to the unit vectors $\hat{\mathbf{u}}_x$ and $\hat{\mathbf{u}}_y$, as shown in Figure 8. Using the latter set of basis vectors, we have the following:

$$^{(j)}\mathbf{u}_i = \begin{bmatrix} \cos\left(^{(j)}\psi_i + ^{(j)}\theta_0\right) \\ \sin\left(^{(j)}\psi_i + ^{(j)}\theta_0\right) \end{bmatrix} \tag{19}$$

With this definition, we can express the change in the point-cloud mean location as follows:

$$^{(2)}\mathbf{x}_S - ^{(1)}\mathbf{x}_S = \frac{^{(2)}g_1\,^{(2)}\mathbf{u}_1 + ^{(2)}g_2\,^{(2)}\mathbf{u}_2}{^{(2)}\theta_U - ^{(2)}\theta_L} - \frac{^{(1)}g_1\,^{(1)}\mathbf{u}_1 + ^{(1)}g_2\,^{(1)}\mathbf{u}_2}{^{(2)}\theta_U - ^{(2)}\theta_L} \tag{20}$$

Again, assuming that the corner feature is fixed in space, the perspective error $\epsilon$ will be opposite the apparent object motion, with $\epsilon = -\left(^{(2)}\mathbf{x}_S - ^{(1)}\mathbf{x}_S\right)$.

To compute Equation (20), we need to evaluate the integral in Equation (16) for $^{(j)}g_i$. Evaluating $^{(j)}g_i$ requires that we integrate between two arbitrary points along a wall segment. The integral is easiest to evaluate if we pin the lower limit of integration at the vertex *O* (where $\theta = 0$) and set the upper limit of integration $\theta_i'$ to align with one of the wall endpoints, which we label $\theta_{we,1}$ and $\theta_{we,2}$. In some cases, however, one wall segment blocks the inner portion of the other wall segment, pushing the lower limit of integration away from the vertex *O* (at an angle $\theta > 0$). To evaluate these cases, we can still use the integral formulation starting at the vertex *O* if we simply difference the integral for the angle associated with the blocking wall segment *j* (out to $\theta_i' = \theta_{we,j}$) from the integral for the angle associated with the blocked wall segment *k* (out to $\theta_i' = \theta_{we,k}$). Because we can construct the integral between any upper and lower limit of integration through such a difference, let us proceed by first evaluating Equation (16) over a range that includes point *O*, with $^{(j)}\Theta_i = \left\{\theta \in [0, ^{(j)}\theta_i']\right\}$. For this case, Equation (16) gives $^{(j)}g_i = ^{(j)}h_i\left(^{(j)}\theta_i'\right)$ with the following:

$$^{(j)}h_i\left(^{(j)}\theta_i'\right) = \begin{cases} -^{(j)}r_0\left(\left|^{(j)}\theta_i'\right|\cos\left(^{(j)}\psi_i\right) + \sin\left(\left|^{(j)}\psi_i\right|\right)\ln\left(\frac{\sin\left(^{(j)}\psi_i - ^{(j)}\theta_i'\right)}{\sin\left(^{(j)}\psi_i\right)}\right)\right) & ^{(j)}\psi_i \neq 0 \\ 0 & ^{(j)}\psi_i = 0 \end{cases}$$

$$\tag{21}$$

The absolute value notation in Equation (21) is redundant when $^{(j)}\psi_i$ and $^{(j)}\theta_i'$ are positive; for the case in which these variables (which always have matching signs) are negative, introducing the absolute value signs gives the correct integral over the range of angles $^{(j)}\Theta_i = \{\theta \in [^{(j)}\theta'_i, 0]\}$. With the absolute value signs, Equation (21) describes the desired integral over a single wall. If only one of the two walls is visible, then $^{(j)}g_i = {}^{(j)}h_i({}^{(j)}\theta'_i)$ for that wall, where $^{(j)}\theta'_i$ is obtained from Equation (14) by setting $^{(j)}f_i = 1$. We will refer to this angle associated with the wall endpoint as $^{(j)}\theta_{we,i}$:

$$^{(j)}\theta_{we,i} = \operatorname{atan}\left(\frac{R\sin\left(^{(j)}\psi_i\right)}{^{(j)}r_0 + R\cos\left(^{(j)}\psi_i\right)}\right) \quad (22)$$

Additional consideration must be given to cases in which both walls are visible or when one of the walls is partly visible behind the other. For these cases, additional logic must be introduced, as summarized in Table 1. To construct this table, we defined the wall angles $^{(j)}\psi_i$ as confined to the range $^{(j)}\psi_i \in (-pi, pi]$, where an angle of zero is aligned with line segment $\overline{LO}$ between the lidar and the vertex O. We also assume that the wall indices are ordered such that the wall closer to the viewer (on either side) is indexed as 1 and the farther wall is indexed as 2. Thus, we have the following:

$$\left|^{(j)}\psi_1\right| > \left|^{(j)}\psi_2\right| \quad (23)$$

The logic in the table describes three possible outcomes. In the first case, if the two angles $^{(j)}\psi_1$ and $^{(j)}\psi_2$ do not have the same sign, then both walls are fully visible, and $^{(j)}g_i = {}^{(j)}h_i({}^{(j)}\theta'_i)$ for each wall. In the last two cases, the walls are on the same side, which causes the rear wall (index 2) to be fully or partially blocked from view by the closer wall (index 1). In both cases, the closer wall analysis is straightforward, with $^{(j)}g_1 = {}^{(j)}h_1({}^{(j)}\theta'_1)$. In the case of full blockage, the second wall does not contribute to the integral: $^{(j)}g_2 = 0$. In the case of a partial blockage, the integral must be computed over just the visible section of the wall, with $^{(j)}g_2 = {}^{(j)}h_2({}^{(j)}\theta_{we,2}) - {}^{(j)}h_2({}^{(j)}\theta_{we,1})$.

**TABLE 1**
Defining Integrals for a Double-Wall Corner in Which Wall 1 is Closer than Wall 2

| Case | Condition | Integral for this condition |
|---|---|---|
| No blockage | $\operatorname{sign}(^{(j)}\psi_1) \neq \operatorname{sign}(^{(j)}\psi_2)$ | $^{(j)}g_1 = {}^{(j)}h_1({}^{(j)}\theta_{we,1})$ |
| | | $^{(j)}g_2 = {}^{(j)}h_2({}^{(j)}\theta_{we,2})$ |
| | | $^{(j)}\theta_U - {}^{(j)}\theta_L = \left|{}^{(j)}\theta_{we,1} - {}^{(j)}\theta_{we,2}\right|$ |
| Blockage | $\operatorname{sign}(^{(j)}\psi_1) = \operatorname{sign}(^{(j)}\psi_2)$ | |
| ↳ Full blockage | Blockage and $\left|{}^{(j)}\theta_{we,1}\right| \geq \left|{}^{(j)}\theta_{we,2}\right|$ | $^{(j)}g_1 = {}^{(j)}h_1({}^{(j)}\theta_{we,1})$ |
| | | $^{(j)}g_2 = 0$ |
| | | $^{(j)}\theta_U - {}^{(j)}\theta_L = \left|{}^{(j)}\theta_{we,1}\right|$ |
| ↳ Partial blockage | Blockage and $\left|{}^{(j)}\theta_{we,1}\right| < \left|{}^{(j)}\theta_{we,2}\right|$ | $^{(j)}g_1 = {}^{(j)}h_1({}^{(j)}\theta_{we,1})$ |
| | | $^{(j)}g_2 = {}^{(j)}h_2({}^{(j)}\theta_{we,2}) - {}^{(j)}h_2({}^{(j)}\theta_{we,1})$ |
| | | $^{(j)}\theta_U - {}^{(j)}\theta_L = \left|{}^{(j)}\theta_{we,2}\right|$ |

## 4 | NUMERICAL RESULTS

To study the perspective error models derived in the previous section, it is helpful to evaluate the equations for a representative scenario. Let us consider the case of a lidar mounted on a mobile platform (e.g., a vehicle or robot) that moves in a straight line past a fixed object.

First, let us consider motion past the cylindrical column. The corresponding geometry is shown in Figure 9. We define the along-track coordinate to be zero when the lidar is closest to the object. We also assume that the lidar first detects the object when it is at an along-track distance of -15$R$ (noting that our simulations express all distances in a nondimensional form, normalized by the column radius $R$). The systematic perspective error $\epsilon$, associated with apparent motion of the point cloud generated from the cylinder, was computed as the negative of Equation (12) and is plotted in Figure 10. Errors were computed as changes in the perceived cylinder location, relative to its initial location at time zero. The starting point is on the left side of the plots shown in Figure 10. When the lidar is closest to the object (at an along-track distance of $x/R = 0$), the distance from the cylinder center in the cross-track direction is set to one of three values: $y/R = \{1, 2, 4\}$. The first case, with $y = R$, is the limiting case in which the lidar just touches the surface of the cylinder. In the subsequent two cases, the cylinder moves progressively farther off the lidar's track.

The along-track error $\epsilon_x$ (Figure 10(a)) and cross-track error $\epsilon_y$ (Figure 10(b)) are shown as functions of the lidar's along-track position $x$. A circular cross-section is shown in the lower plot, as a reminder that the plot describes a cylinder shifted to the left (positive $y$-direction) of the lidar. In all cases, the along-track error $\epsilon_x$ becomes increasingly negative as the lidar moves from left to right. The increase in the error per unit of vehicle motion is a function of the absolute distance of the vehicle from $x/R = 0$ (where the vehicle is adjacent to the cylinder). This trend explains the reflection symmetry of the error curve about this point. The negative along-track errors arise because the point cloud rotates around the cylinder from its left (more negative) side to its right (more positive) side. Because the point-cloud position moves to the right (positive), the vehicle measures less than the true distance of travel (a negative error in the along-track direction) if the cylindrical column is assumed to be stationary. The accumulated error $\epsilon_x/R$ is approximately $\pi/2$ (which is reasonable considering that the point-cloud center moves from approximately $-\pi/4$ initially to $+\pi/4$ in the end, as expected for the limit case shown in Figure 7).

In contrast to the along-track error $\epsilon_x$, which grows in magnitude over time, the cross-track error $\epsilon_y$ grows to a maximum at the time point when the object is viewed laterally (at $x/R=0$). As the lidar travels past the object, the error decreases back to zero. This trend makes sense. Initially, when viewed from the side, the cross-track position of the point cloud is near the cross-track position of the cylinder's center. As the lidar moves past the object, the point cloud shifts around the circular cross-section until it is centered on the side closest to the lidar at $x/R=0$. This represents a negative apparent motion for the point cloud; thus, the vehicle appears to move in the positive direction ($\epsilon_y > 0$). Later, as the lidar moves further to the right, the point cloud continues to rotate around the circular cross section, toward the column's right side. This represents a positive change in the point cloud's y coordinate, back toward its original value (with the lateral error $\epsilon_y$ also returning to zero as the lidar moves off the right edge of Figure 10(b)).

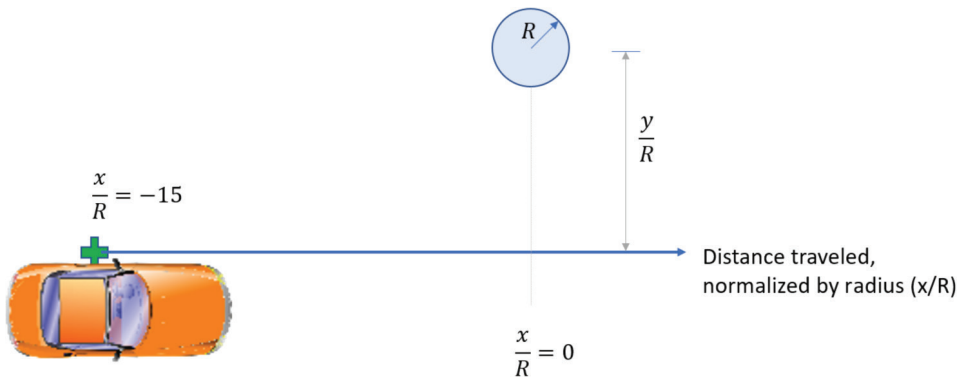

**FIGURE 9** Schematic for a lidar-equipped vehicle moving past a cylindrical object

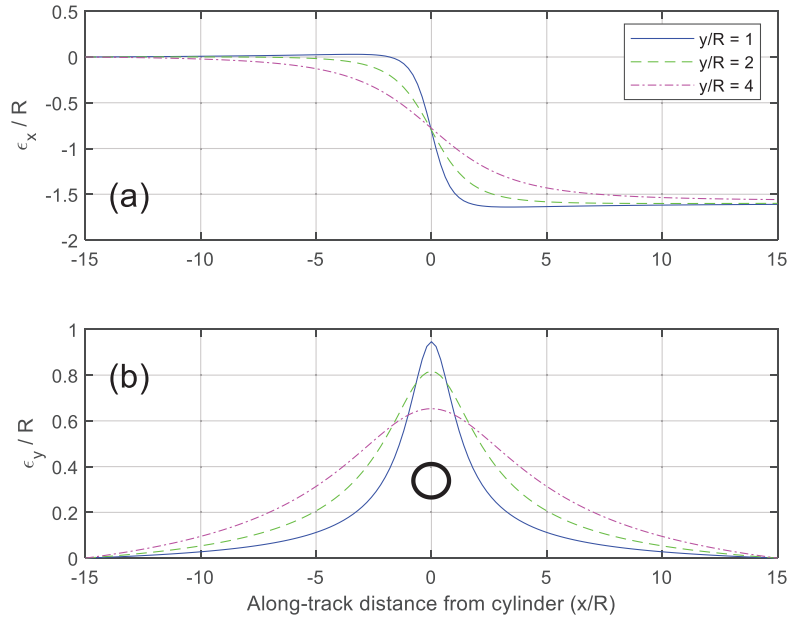

**FIGURE 10** Perspective error that results when a lidar moves on a straight trajectory past a cylindrical column

As the lidar trajectory is shifted increasingly far from the object (going from $y=R$ to $y=4R$), the changes are more gradual. For instance, greater lateral spacing y causes the along-track error $\epsilon_x$ to accumulate sooner and the slope to be more gradual (but still resulting in roughly the same level of total error $\epsilon_x$, as shown on the right side of Figure 10). For the cross-track error, the error curve is wider (with error triggering sooner) and shorter (lower maximum lateral error) for higher values of y.

The trends observed for a dual-wall corner with a convex configuration are similar to the trends for the column (see Figure 11(a) and Figure 11(b), noting that the latter is marked with a "V-shaped" corner). In this configuration, the wall segments are each canted 45° back from the lidar's line of travel. Only the outer faces of the wall segments are seen as the lidar passes. The along-track error $\epsilon_x$ increases monotonically in magnitude as the point-cloud mean moves from the negative wall to the positive wall. The cross-track error $\epsilon_y$ increases and then decreases, with the lateral error peaking when the lidar is closest to the vertex (at $x/R=0$). The only detail that is notably different between the cylinder and the convex dual-wall corner is the peak error magnitude, noting that the objects have cross-sections of similar size (cylindrical column of radius R and dual-wall corner for which each edge is length R). For the convex corner in Figure 11(a), the largest along-track

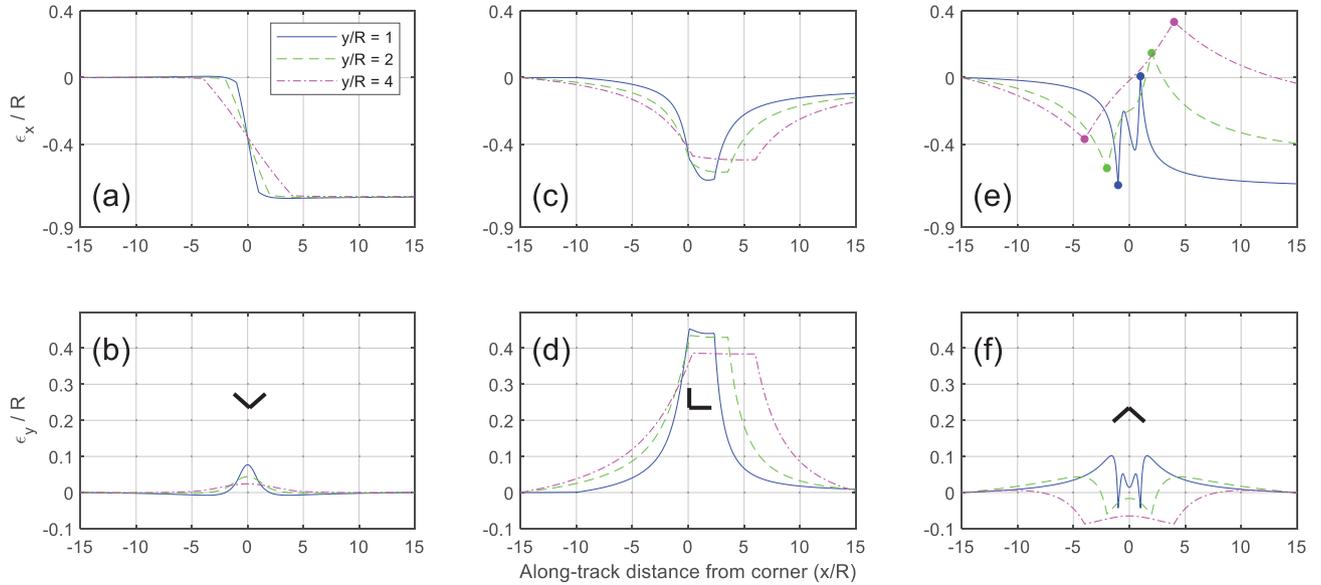

**FIGURE 11** Perspective error relative to the starting position (at $x/R = -15$) as the lidar moves in a straight line past a corner

Each column shows a corner with a different orientation (as depicted in the lower row). In (a) and (b), the corner opens away from the lidar (convex when viewed from $x = 0$); in (c) and (d), the corner opens to the positive x and y directions (flat when viewed from $x = 0$); in (e) and (f), the corner opens toward the lidar (concave when viewed from $x = 0$).

error is approximately $R/\sqrt{2}$ in magnitude (the distance between the midpoints of the two walls), whereas the largest cross-track error is a very small fraction of $R$ (less than $0.1R$) because both walls are entirely visible at $x/R = 0$. As the location of the walls shifts farther laterally (vertex $y/R$ increases from 1 to 4), the trends again become more gradual.

The error profile differs somewhat for other orientations of the dual-wall corner. Whereas a convex corner is a reasonable approximation for the cylinder, the L-shaped corner (Figure 11(c) and Figure 11(d)) and the concave corner (Figure 11(e) and Figure 11(f)) exhibit unique properties related to occlusions of one wall by the other. All visible wall segments are considered in our analysis here; however, we acknowledge that the more distant wall segment might be excluded in a practical implementation because of shadow mitigation (McDermott & Rife, 2022a).

Let us consider the L-shaped corner (Figure 11(c) and Figure 11(d)). In this case, the corner has one wall aligned with the *y*-direction and one wall aligned with the *x*-direction, with the vertex on the side closest to the lidar as it passes. Initially, the lidar primarily sees only the outside *y*-aligned wall. This wall disappears when the lidar reaches $x/R = 0$, at which point only the *x*-aligned wall is visible. Eventually the inside of the *y*-aligned wall becomes visible again, and by the time the lidar reaches its final location $(x/R = 15)$, this *y*-aligned wall dominates the point cloud. Because the lidar starts and finishes primarily seeing opposite sides of the *y*-aligned wall, which has no thickness, the net $\epsilon_x$ is nearly zero. The $\epsilon_y$ error climbs to a maximum of $0.45R$ (for the vertex at $y/R = 1$) when the lidar passes the corner, which is reasonable because the lidar sees only the *x*-aligned wall at this point, which is closer than the *y*-aligned wall by a distance of $0.5R$.

Now let us consider the concave corner (Figure 11(e) and Figure 11(f)). In this case, the corner opens toward the lidar, such that both walls are canted 45° toward the lidar's track. As the lidar travels, it nearly always sees at least a fraction of both walls. Due to a combination of point-density effects (see Figure 5) and switching between viewing the inner and outer faces of each wall, the point-cloud mean moves back and forth as the lidar progresses. As a result, the $\epsilon_x$ and $\epsilon_y$ errors increase and decrease several times. Although a substantial $\epsilon_x$ error accumulates, the $\epsilon_y$ error generally remains in a small band near zero (smaller in magnitude than approximately $0.1R$). Interestingly, the sign of the net $\epsilon_x$ error is positive in the region where only the interior faces of the convex corner are visible. (Exterior faces are not visible between the dot markers, shown in Figure 11(e)). The positive net $\epsilon_x$ indicates an overprediction of the distance traveled. Thus, when only interior walls are visible, the point cloud generally seems to move backward, which causes a perspective error that increases the lidar's apparent travel distance.

It is important to note that the dual-wall corner was modeled as a free-standing object, visible from all sides. If the concave corner were embedded in a larger wall (e.g., an inverted V-shape notched into a flat wall), then only the interior faces of the corner would be visible, much like the case with $y/R = 4$. By extension, the perspective error for the notched wall would also result in an overestimation of the distance traveled. As a generalization, we can state that perspective errors on convex objects tend to cause an underestimation of the distance traveled, whereas perspective errors on concave objects tend to cause an overestimation.

To translate these results into real-world terms, consider the cylinder case shown in Figure 10(a), where the along-track error accumulates to reach a maximum of $1.5R$. For the case of a typical telephone pole (approximately 20-cm radius near the base of the pole), this corresponds to an accumulated bias error of approximately 3 m in one voxel. If similar errors occur in 10% of all voxels and if the voxels are weighted similarly (noting that weights are related to predicted random-error covariances for each voxel), then the combined biases would result in roughly a 0.3-m bias in the relative position solution. We observe that this level of errors is in the ballpark of the experimental errors shown in Figure 1. Importantly, this level of bias is quite large compared with random-walk errors for scan matching, which implies that some form of perspective error mitigation is necessary to define a voxel-based scan-matching approach that delivers both high accuracy and high integrity.

## 5 | MITIGATION

Given that perspective errors can significantly bias lidar scan matching in some voxels, it is desirable to mitigate these errors in order to increase the accuracy. This section considers several approaches for mitigating perspective error. Investigating the effectiveness of these various techniques is an important topic for future work.

### 5.1 | Voxel Dimensions

Perhaps the most straightforward approach for mitigating perspective errors is to use smaller voxels. If smaller voxels are used, then missing-data regions (e.g., the red arcs in Figure 4) can ideally be confined within a small voxel. Voxel processing is only conducted when a sufficient number of samples are measured within the voxel for both signals (McDermott & Rife, 2022a); thus, voxels are not processed if lidar samples are present from only one of two scans. To obtain insight into the trends between the worst case (when the object is entirely contained within one voxel, as analyzed above) and the best case (when voxels are small enough that their boundaries align with the boundaries of the missing-data regions), let us consider the introduction of progressively smaller voxels, as illustrated in Figure 12. This figure shows four gridding arrangements superimposed on a cylindrical column. The analysis in the prior section can be applied when the grid cell is large enough to fully contain the cylinder; for that case, the point-cloud mean can fall on any point in an annulus (see Figure 12(a)). The annulus is bounded on the outside by the outer column boundary and on the inside by a ring at $\frac{\pi}{4}R$ (which is the asymptote of the point-cloud mean shown in Figure 7). Within the annulus, the specific location of the point-cloud mean depends on the location of the lidar. For the case in which the lidar circles halfway around the column, near the column surface, the point-cloud mean will shift by the column diameter (by a distance of 2$R$). In the worst case, the voxel boundary (dashed line) closely matches the size of the cylindrical column (Figure 12(a)), resulting in a perspective error that is equal in size to the voxel.

If the voxel edge length is cut in half, from 2$R$ to $R$ (as in Figure 12(b)), then the analysis of the prior sections no longer describes the point-cloud mean. However, we can still limit the point-cloud mean to the red area shown in Figure 12(b) by noting that the convex hull of the surface can be as large as a quarter circle and that the mean of points on that surface must lie within the convex hull (Boyd et al., 2004). The shortest dimension of the convex hull lies along the axis extending from the center of the circular cross-section. It is straightforward to show that the depth of the convex hull in this radial direction is $\Delta_r = R(1 - \cos\phi_h)$, where $\phi_h$ is half the angle between the radial spokes extending from the circle center to the sharp endpoints of the convex hull. The width of the convex hull in the perpendicular, or circumferential, direction is $\Delta_\phi = 2R\sin\phi_h$. In the worst-case alignment of the cylinder relative to the voxel boundaries, the circular arc connects the voxel corners (as shown in Figure 12(b)), such that $\Delta_\phi = \sqrt{2}R$ (or 1.41$R$) and $\Delta_r = 0.29R$. In short, halving the voxel width reduces the worst error by $\sqrt{2}$.

This trend continues as the voxels decrease further. When there are three voxels across the diameter of the cylinder (shown in Figure 12(c); edge width of $\frac{2}{3}R$), the worst case error is $\Delta_\phi = \frac{2\sqrt{2}}{3}R = 0.94R$. When there are four voxels across the diameter of the cylinder (shown in Figure 12(d); edge width of $\frac{1}{2}R$), the worst-case error is $\Delta_\phi = \frac{\sqrt{2}}{2}R = 0.71R$.

A secondary benefit of using smaller voxels is that fewer voxels align in the worst-case configuration as the voxels become smaller. In Figure 12(b), it is possible for all four voxels to align in the worst-case configuration. In Figure 12(c), it is possible for only one of nine voxels shown to align in the worst-case configuration.

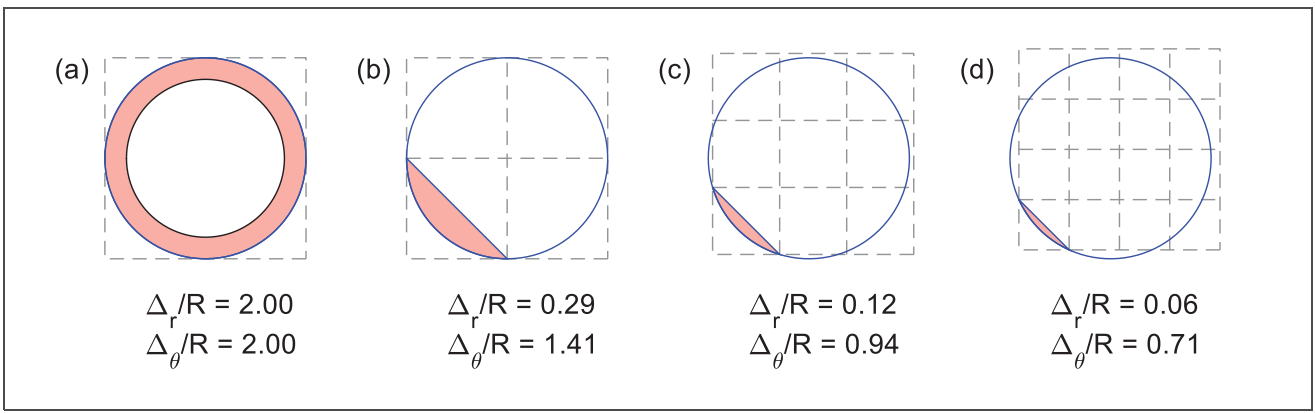

FIGURE 12  Effect of voxel size on worst-case perspective error

In Figure 12(d), it is possible for only one of sixteen voxels shown to align in the worst-case configuration. In sum, a smaller voxel size always reduces the magnitude of the worst-case error and, at least in the case shown, also reduces the number of voxels that experience the worst-case error. Although a higher voxel count can mitigate perspective error, the additional voxels increase the computational complexity; consequently, adaptive resolution methods may offer benefits in balancing accuracy and computational cost (Eckart et al., 2018).

## 5.2 | Exclusion of Lateral Features

Another possible mitigation approach is to avoid using nearby voxels orthogonal to the lidar velocity, at least if they contain a compact feature. This heuristic approach is driven by the observation that perspective errors accumulate most rapidly when a feature is approximately orthogonal to the direction of motion (near $x/R = 0$, as shown in Figure 10 and Figure 11). As the figures show, the perspective error is most pronounced when the feature is close to the lidar (when $y/R$ is low).

In large part, the perspective error accumulates rapidly because the viewing angle, from a moving observer to a point $O$, changes quickly when the point is positioned orthogonal to the direction of motion at a relatively short distance. We can see this by defining two position vectors, $^{(1)}\mathbf{x}_o$ and $^{(2)}\mathbf{x}_o$, which describe the vector to a point $O$ from two viewing locations, labeled 1 and 2. The change in viewing angle $\phi$ is the angle between these vectors, which we can compute with the dot product:

$$\phi = \mathrm{acos}\left(\frac{^{(2)}\mathbf{x}_o \cdot {^{(1)}\mathbf{x}_o}}{\left\|^{(2)}\mathbf{x}_o\right\|\left\|^{(1)}\mathbf{x}_o\right\|}\right) \qquad (24)$$

Let us assume that the motion carries the viewer a distance $d$ along the direction aligned with the $x$-axis. Then, the two vectors are related by $^{(2)}\mathbf{x}_o = \hat{\mathbf{u}}_x d + {^{(1)}\mathbf{x}_o}$. We can now use Equation (24) to compute the viewing angle $\phi$ for points located at any position $(x,y)$ relative to the midpoint of the two viewing locations. A contour plot of the viewing angle for objects at various locations is plotted in Figure 13, where the horizontal ($x$-axis) and vertical ($y$-axis) coordinates correspond to the position of point $O$ and where the isocontours describe the angle shift (in degrees) for a small horizontal movement $d$. Note that the figure describes the location of point $O$ in distances normalized by $d$, such that the figure axes are nondimensional.

The isocontours show that the largest changes in viewing angle (in degrees) are along the vertical axis through $x/R = 0$ (orthogonal to the motion) and closer to the observer. The trends of large changes in viewing angle shown in Figure 13 are consistent with the trends of high perspective error shown in Figure 10 and Figure 11, supporting the correlation between these two parameters.

An important caveat is that perspective errors only accumulate rapidly for surfaces that contain significant curvature within a cell. Changes in viewing angle allow the viewer to "see around the bend" for compact objects such as cylinders and dual-wall corners, with motion revealing a new region of the object's surface. For flat planes, the full surface remains in view as long as the viewer stays in front of the plane; in this case, perspective errors can result only from density changes (see Figure 5), a relatively minor effect. Even for mildly curved surfaces, the error perpendicular to the surface is small (noting, for example, the low values of the perpendicular error $\Delta_R/R$ for short circular arcs, as shown in Figure 12(c) and Figure 12(d)). By contrast, for corners and objects with large curvature within a voxel, the change in viewing angle reveals new portions of the object surface as the lidar moves.

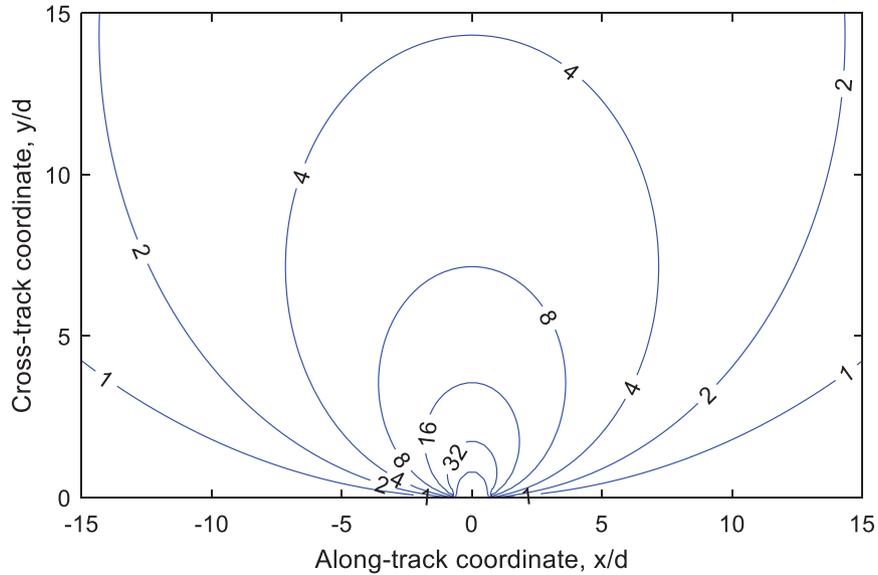

**FIGURE 13** Change in viewing angle (degrees) as a function of object location for movement in the *x*-direction of distance *d*

The distinction between compact and extended surfaces is a major feature of the ICET algorithm, developed by McDermott and Rife (2022b). It may be useful to leverage this algorithm's ability to distinguish compact and extended features in order to develop a future strategy for mitigating perspective errors.

### 5.3 | Monitoring for Perspective Change

Another possible mitigation approach for perspective errors is to introduce a monitor to assess for significant changes in the distribution of lidar points within a given voxel. A change in distribution shape may be a good indicator that a new region of a surface has appeared in view, a condition that is necessary for large perspective errors. The design of such a monitor is outside the scope of this paper.

### 5.4 | Object Reconstruction

The basic tenet of scan matching is that commonality should be identified between two lidar scans so that the scans can be aligned. In concept, it may be possible to extract more information from a scan by classifying objects. If an object can be identified as the same feature in two scans, even if it is viewed from radically different angles, then that feature could be used as a landmark for localization. There are two challenging problems here: reliably identifying landmarks and estimating their associated boundaries to register different views of a landmark (Zhou & Tuzel, 2018; Li & Wang, 2020; Xu et al., 2020). These problems are being studied in the ML community, but developing a rigorous assurance case for ML-based algorithms remains a hurdle for safety-of-life applications.

# CONCLUSION

The main goal of this paper was to characterize lidar perspective errors, a significant source of systematic error in lidar scan matching and odometry. In particular, we introduced an analytic model of perspective errors for voxel-based scan-matching algorithms, such as NDT and ICET, which attempt to match the mean of the point distribution within a voxel. We showed that the systematic effect of the perspective error can accumulate to a significant value, as large as the diameter of a cylindrical column, for instance. For convex objects (e.g., columns, poles, or outer corners), the perspective error underpredicts the lidar's forward motion. For concave objects (e.g., inner corners), an overprediction of forward motion is possible. Both convex and concave objects may result in a false inference of lateral motion, but these effects are transient and disappear as the lidar continues on its path. Concepts were discussed for the mitigation of perspective errors, including the use of smaller voxels, lateral-feature exclusion, monitoring, and object reconstruction.

## ACKNOWLEDGMENTS

The authors wish to acknowledge and thank the U.S. Department of Transportation (DOT) Joint Program Office and the Office of the Assistant Secretary for Research and Technology for sponsorship of this work. We also gratefully acknowledge National Science Foundation (NSF) grant CNS-1836942, which supported specific aspects of this research. The opinions discussed here are those of the authors and do not necessarily represent those of the DOT, NSF, or other affiliated agencies.